\documentclass[11pt,a4paper]{article}
\usepackage[hyperref]{eacl2021}
\usepackage{times}
\usepackage{latexsym}

\usepackage{microtype}

\aclfinalcopy 
\def\aclpaperid{451} 

\usepackage{url}
\usepackage{multirow,booktabs,makecell}
\usepackage{booktabs}
\usepackage{graphicx}
\usepackage{adjustbox}
\usepackage{color}
\usepackage{comment}
\usepackage{subcaption}
\usepackage[T1]{fontenc}
\usepackage{enumitem}

\usepackage{amsmath,amsfonts,bm}

\def\eqref#1{equation~\ref{#1}}

\def\1{\bm{1}}

\def\rmI{{\mathbf{I}}}

\def\vh{{\bm{h}}}

\def\vs{{\bm{s}}}

\def\vv{{\bm{v}}}
\def\vw{{\bm{w}}}

\def\evalpha{{\alpha}}

\def\mW{{\bm{W}}}

\DeclareMathAlphabet{\mathsfit}{\encodingdefault}{\sfdefault}{m}{sl}
\SetMathAlphabet{\mathsfit}{bold}{\encodingdefault}{\sfdefault}{bx}{n}

\def\gA{{\mathcal{A}}}
\def\gB{{\mathcal{B}}}
\def\gC{{\mathcal{C}}}

\def\gE{{\mathcal{E}}}
\def\gF{{\mathcal{F}}}

\def\gR{{\mathcal{R}}}

\def\gT{{\mathcal{T}}}

\def\gX{{\mathcal{X}}}

\newcommand{\R}{\mathbb{R}}

\title{Multimodal Text Style Transfer for\\ Outdoor Vision-and-Language Navigation}

\author{Wanrong Zhu\textsuperscript{\dag}, Xin Eric Wang\textsuperscript{\ddag}, Tsu-Jui Fu\textsuperscript{\dag}, An Yan\textsuperscript{\S}, 
\\ \textbf{Pradyumna Narayana\textsuperscript{*}, 
Kazoo Sone\textsuperscript{*}, Sugato Basu\textsuperscript{*}, William Yang Wang\textsuperscript{\dag}} \\
\textsuperscript{\dag}UC Santa Barbara, 
\textsuperscript{\ddag}UC Santa Cruz,
\textsuperscript{\S}UC San Diego,
\textsuperscript{*}Google \\
\texttt{\small \{wanrongzhu,tsu-juifu,william\}@cs.ucsb.edu}, 
\texttt{\small xwang366@ucsc.edu},\\
\texttt{\small ayan@eng.ucsd.edu},
\texttt{\small \{pradyn,sone,sugato\}@google.com}
}

\date{}

\begin{document}
\maketitle
\begin{abstract}

One of the most challenging topics in Natural Language Processing (NLP) is visually-grounded language understanding and reasoning. Outdoor vision-and-language navigation (VLN) is such a task where an agent follows natural language instructions and navigates a real-life urban environment. Due to the lack of human-annotated instructions that illustrate  intricate urban scenes, outdoor VLN remains a challenging task to solve.
This paper introduces a Multimodal Text Style Transfer (MTST) learning approach and leverages external multimodal resources to mitigate data scarcity in outdoor navigation tasks. 
We first enrich the navigation data by transferring the style of the instructions generated by Google Maps API, then pre-train the navigator with the augmented external outdoor navigation dataset.
Experimental results show that our MTST learning approach is model-agnostic, and our MTST approach significantly outperforms the baseline models on the outdoor VLN task, improving task completion rate by 8.7\% relatively on the test set. 
\footnote{Our code and dataset is released at \url{https://github.com/VegB/VLN-Transformer}.}
\end{abstract}

\section{Introduction}

\begin{figure}[t]
    \centering
    \includegraphics[width=\linewidth]{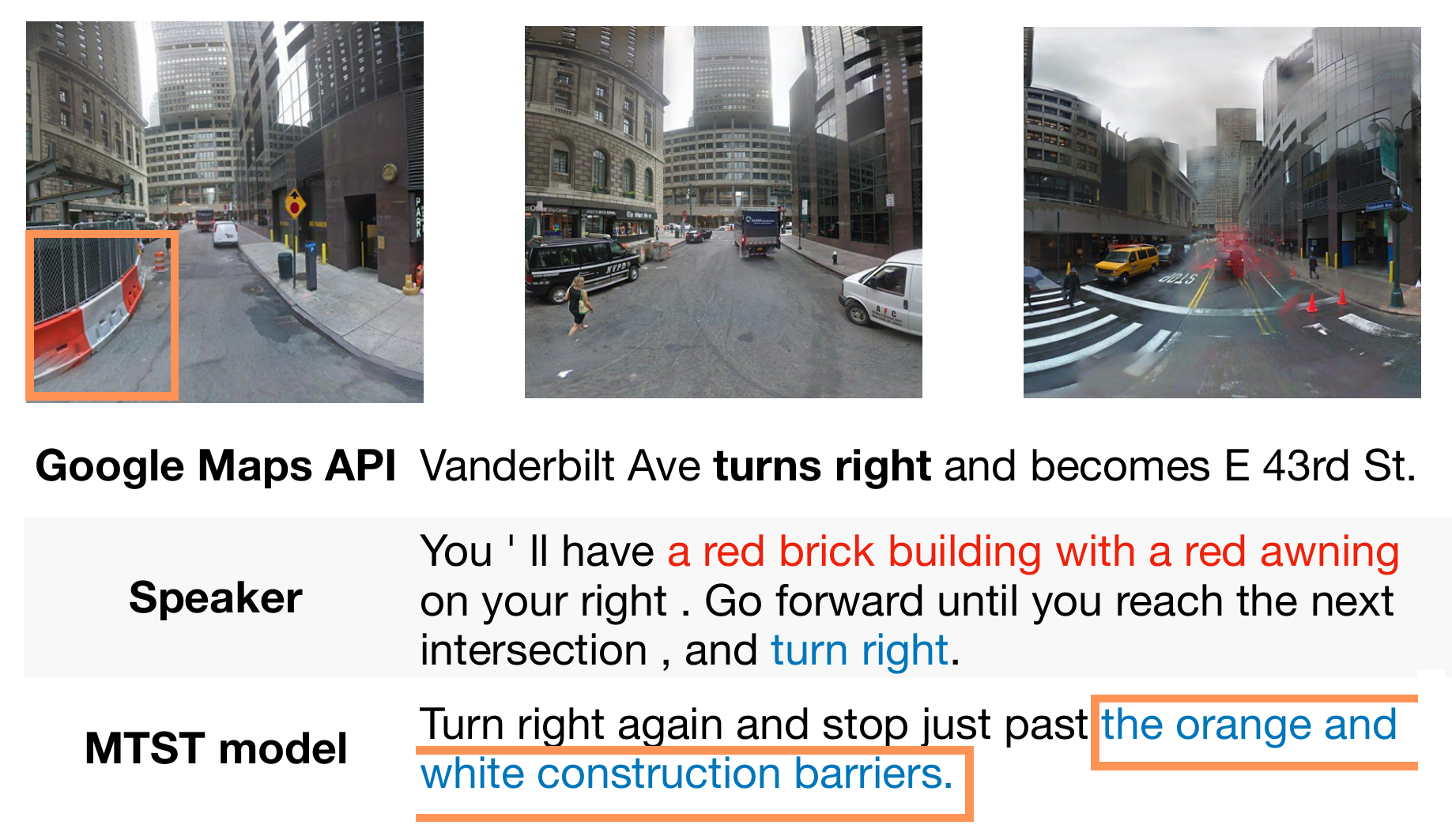}
    \caption{An outdoor VLN example with instructions generated by Google Maps API (ground truth), the Speaker model, and our MTST model. Tokens marked in red indicate incorrectly generated instructions, while the blue tokens suggest alignments with the ground truth. The orange bounding boxes show that the objects in the surrounding environment have been successfully injected into the style-modified instruction.}
    \label{fig:style_transfer_example}
\end{figure}

A key challenge for Artificial Intelligence research is to go beyond static observational data and consider more challenging settings that involve dynamic actions and incremental decision-making processes~\cite{pearl2018book}.
Outdoor vision-and-language navigation (VLN) is such a task, where an agent navigates in an urban environment by grounding natural language instructions in visual scenes, as illustrated in Fig.~\ref{fig:style_transfer_example}. 
To generate a series of correct actions, the navigation agent must comprehend the instructions and reason through the visual environment. 

Different from indoor navigation ~\cite{anderson2018r2r,Wang-2018,fried2018speakerfollower,wang2019reinforced,ma2019self,tan2019learning,ma2019regretful,ke2019tactical}, the outdoor navigation task takes place in urban environments that contain diverse street views~\cite{mirowski2018learning,chen2019touchdown,mehta2020retouchdown}. 
The vast urban area leads to a much larger space for an agent to explore and usually contains longer trajectories and a wider range of objects for visual grounding. This requires more informative instructions to address the complex navigation environment. 
However, it is expensive to collect human-annotated instructions that depict the complicated visual scenes to train a navigation agent. The issue of data scarcity limits the navigator's performance in the outdoor VLN task.

To deal with the data scarcity issue, ~\citet{fried2018speakerfollower} proposes a Speaker model to generate additional training pairs. 
However, synthesizing instructions purely from visual signals is hard, especially for outdoor environments, due to visual complexity.
On the other hand, template-based navigation instructions on the street view can be easily obtained via the Google Map API, which may serve as additional learning signals to boost outdoor navigation tasks. 
But instructions generated by Google Maps API mainly consist of street names and directions, while human-annotated instructions in the outdoor navigation task frequently refer to street-view objects in the panorama. 
The distinct instruction style hinders the full utilization of external resources.

Therefore, we present a novel Multimodal Text Style Transfer (MTST) learning approach to narrow the gap between template-based instructions in the external resources and the human-annotated instructions for the outdoor navigation task. It can infer style-modified instructions for trajectories in the external resources and thus mitigate the data scarcity issue.
Our approach can inject more visual objects in the navigation environment to the instructions (Fig.~\ref{fig:style_transfer_example}), while providing direction guidance. The enriched object-related information can help the navigation agent learn the grounding between the visual environment and the instruction.

Moreover, different from previous LSTM-based navigation agents, we propose a new VLN Transformer to predict outdoor navigation actions.
Experimental results show that utilizing external resources provided by Google Maps API during the pre-training process improves the navigation agent's performance on Touchdown, a dataset for outdoor VLN~\cite{chen2019touchdown}.
In addition, pre-training with the style-modified instructions generated by our multimodal text style transfer model can further improve navigation performance and make the pre-training process more robust. In summary, the contribution of our work is four-fold:
\begin{itemize}
    \item We present a new Multimodal Text Style Transfer learning approach to generate style-modified instructions for external resources and tackle the data scarcity issue in the outdoor VLN task. 
    \item We provide the Manh-50 dataset with style-modified instructions as an auxiliary dataset for outdoor VLN training. 
    \item We propose a novel VLN Transformer model as the navigation agent for outdoor VLN and validate its effectiveness.  
    \item We improve the task completion rate by 8.7\% relatively on the test set for the outdoor VLN task with the VLN Transformer model pre-trained on the external resources processed by our MTST approach. 
\end{itemize}

\section{Related Work}
\noindent\textbf{Vision-and-Language Navigation (VLN)~} is a task that requires an agent to achieve the final goal based on the given instructions in a 3D environment. 
Besides the generalizability problem studied by previous works~\cite{Wang-2018,wang2019reinforced,tan2019learning,zhang2020diagnosing}, the data scarcity problem is another critical issue for the VLN task, expecially in the outdoor environment\cite{chen2019touchdown,mehta2020retouchdown,DBLP:conf/emnlp/Xiang0W20}. ~\citet{fried2018speakerfollower} obtains a broad set of augmented training data for VLN by sampling trajectories in the navigation environment and using the Speaker model to back-translate their instructions. However, the Speaker model might cause the error propagation issue since it is not trained on large corpora to optimize generalization.
While most existing works select navigation actions dynamically along the way in the unseen environment during testing, \citet{majumdar2020improving} proposes to test in previously explored environments and convert the VLN task to a classification task over the possible paths. This approach performs well in the indoor setting, but is not suitable for outdoor VLN where the environment graph is different.

\noindent\textbf{Multimodal Pre-training ~} has attracted much attention to improving multimodal tasks performances. The models usually adopt the Transformer structure to encode the visual features and the textual features \cite{Tan2019LXMERTLC, Lu2019ViLBERTPT,chen2019uniter, Sun2019VideoBERTAJ, Li2019VisualBERTAS, huang2020pixelbert, Luo2020UniViLMAU, Li2020UnicoderVLAU,DBLP:conf/acl/ZhengGK20,DBLP:conf/cvpr/WeiZLZW20,DBLP:conf/acl/TsaiBLKMS19}.
During pre-training, these models use tasks such as masked language modeling, masked region modeling, image-text matching to learn the cross-modal encoding ability, which later benefits the multimodal downstream tasks.
\citet{majumdar2020improving} proposes to use image-text pairs from the web to pre-train VLN-BERT, a visiolinguistic transformer-based model similar to the model proposed by \citet{Lu2019ViLBERTPT}.

A concurrent work by~\citet{DBLP:conf/cvpr/HaoLLCG20} proposes to use Transformer for indoor VLN. Our VLN Transformer is different from their model in several key aspects:
(1) The pre-training objectives are different: ~\citet{DBLP:conf/cvpr/HaoLLCG20} pre-trains the model on the same dataset for training, while we create an augmented, stylized dataset for outdoor VLN using the proposed MTST method.  
(2) Benefiting from the effective external resource, a simple navigation loss is employed in our VLN Transformer, while they adopt the masked language modeling to better train their model. 
(3) Model-wise, instead of encoding the whole instruction into one feature, we use sentence-level encoding since Touchdown instructions are much longer than R2R instructions.
(4) We encode the trajectory history, while their model encodes the panorama for the current step.

\begin{figure*}[t]
\centering
\includegraphics[width=0.9\textwidth]{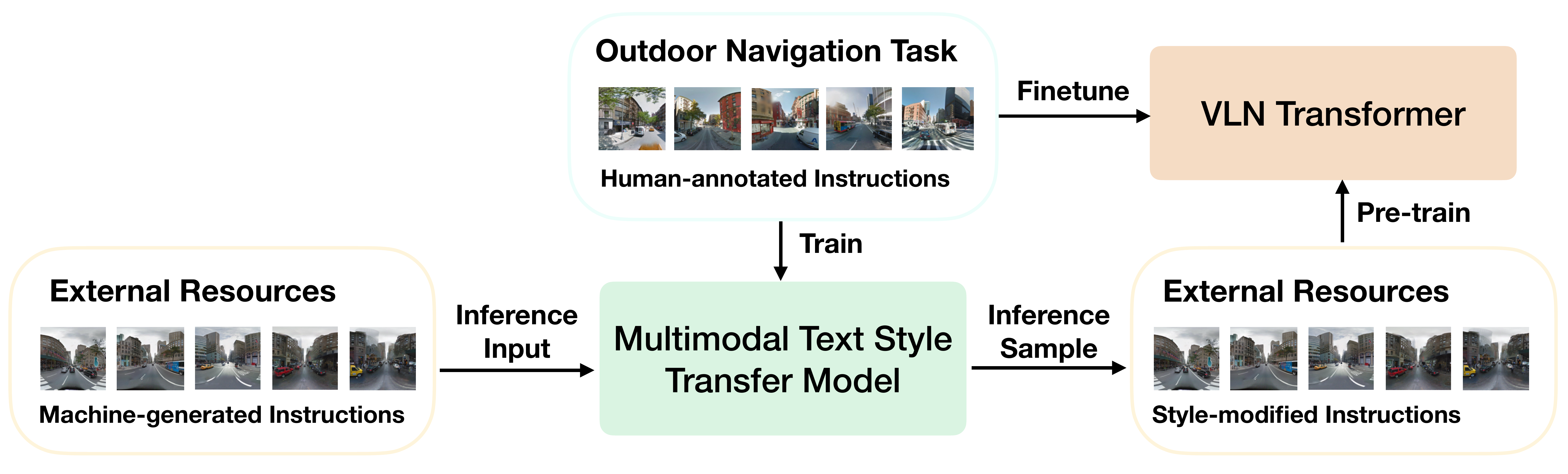}
\caption{An overview of the Multimodal Text Style Transfer (MTST) learning approach for vision-and-language navigation in real-life urban environments. Details are described in Section ~\ref{sec:overview}.}
\label{fig:process_overview}
\end{figure*}

\noindent\textbf{Unsupervised Text Style Transfer~} is an approach to mitigate the lack of parallel data for supervised training. One line of work encodes the text into a latent vector and manipulate the text representation in the latent space to transfer the style. \citet{Shen2017StyleTF, Hu2017TowardCG, Yang2018UnsupervisedTS} use variational auto-encoder to encode the text, and use a discriminator to modify text style. \citet{John2018DisentangledRL, Fu2018StyleTI} rely on models with encoder-decoder structure to transfer the style.
Another line of work enriches the training data by generating pseudo-parallel data via back-translation
~\cite{Artetxe2018UnsupervisedNM, Lample2018PhraseBasedN, Lample2018UnsupervisedMT, Zhang2018StyleTA}.


\section{Methods}

\subsection{Task Definition}

In the vision-and-language navigation task, the reasoning navigator is asked to find the correct path to reach the target location following the instructions (a set of sentences) $\gX = \{s_1, s_2, \dots, s_m\}$.
The navigation procedure can be viewed as a series of decision making processes. At each time step $t$, the navigation environment presents an image view $\vv_t$. With reference to the instruction $\gX$ and the visual view $\vv_t$, the navigator is expected to choose an action $a_t \in \gA$. The action set $\gA$ for urban environment navigation usually contains four actions, namely \textit{turn left}, \textit{turn right}, \textit{go forward}, and \textit{stop}.

\subsection{Overview}
\label{sec:overview}
Our Multimodal Text Style Transfer (MTST) learning mainly consists of two modules, namely the \emph{multimodal text style transfer model} and the \emph{VLN Transformer}. Fig.~\ref{fig:process_overview} provides an overview of our MTST approach. 
We use the multimodal text style transfer model to narrow the gap between the human-annotated instructions for the outdoor navigation task and the machine-generated instructions in the external resources. The multimodal text style transfer model is trained on the dataset for outdoor navigation, and it learns to infer style-modified instructions for trajectories in the external resources. 
The VLN Transformer is the navigation agent that generates actions for the outdoor VLN task. It is trained with a two-stage training pipeline. We first pre-train the VLN Transformer on the external resources with the style-modified instructions and then fine-tune it on the outdoor navigation dataset.

\subsection{Multimodal Text Style Transfer Model}
\label{sec:mm_text_style_transfer}

\begin{table}
\begin{adjustbox}{width=\linewidth,center}
\begin{tabular}{c c }
\cmidrule[\heavyrulewidth]{1-2}
\textbf{Source} & \textbf{Instruction} \\ \cmidrule{1-2}
Google Maps API     & \makecell[l]{Head northwest on E 23rd St toward 2nd Ave.\\ Turn left at the 2nd cross street onto 3rd Ave. }\\ \cmidrule{1-2}
Human Annotator     & \makecell[l]{Orient yourself so you are facing the same as\\the traffic on the 4 lane road. Travel down this\\road until the first intersection. Turn left and go\\down this street with the flow of traffic. You'll\\see a black and white stripped awning on your\\right as you travel down the street.} \\ \cmidrule[\heavyrulewidth]{1-2}
\end{tabular}
\end{adjustbox}
\caption{For the outdoor VLN task, the instructions provided by Google Maps API is distinct from the instructions written by human annotators.}
\label{tab:instr_comparison}

\end{table}

\noindent\textbf{Instruction Style~}
\label{instr_style}
The navigation instructions vary across different outdoor VLN datasets. 
As shown in Table \ref{tab:instr_comparison}, the instructions generated by Google Maps API is template-based and mainly consists of street names and directions.
In contrast, human-annotated instructions for the outdoor VLN task emphasize the visual environment's attributes as navigation targets. It frequently refers to objects in the panorama, such as traffic lights, cars, awnings, etc. 
The goal of conducting multimodal text style transfer is to inject more object-related information in the surrounding navigation environment to the machine-generated instruction while keeping the correct guiding signals.

\begin{figure*}[t]
\centering
\includegraphics[width=\linewidth]{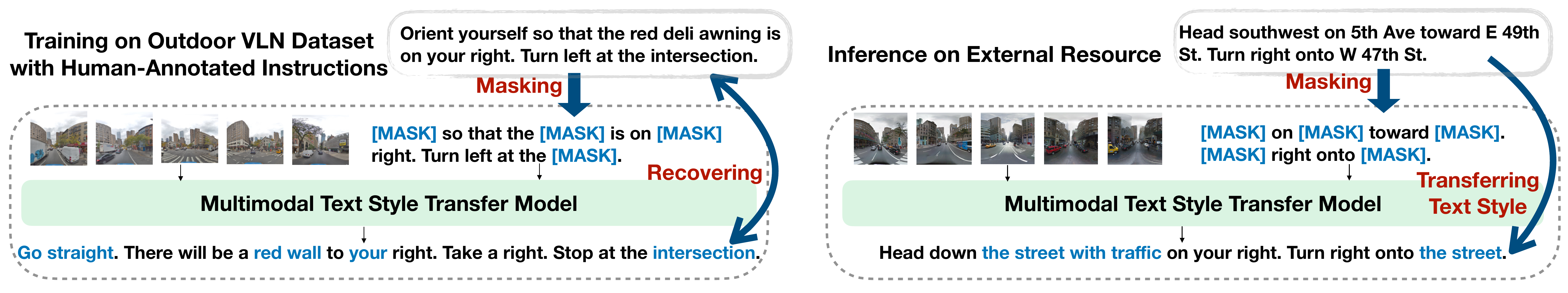}
\caption{An example of the training and inference process of the multimodal text style transfer model. During training, we mask out the objects in the human-annotated instructions to get the instruction template. The model takes both the trajectory and the instruction skeleton as input, and the training objective is to recover the instructions with objects. When inferring new instructions for external trajectories, we mask the street names in the original instructions and prompt the model to generate new object-grounded instructions. }
\label{fig:style_transfer_process}
\end{figure*}

\noindent\textbf{Masking-and-Recovering Scheme~}
The multimodal text style transfer model is trained with a ``masking-and-recovering"~\cite{zhu2019text,Liu2019TIGSAI,Donahue2020EnablingLM,Huang2020INSETSI} scheme to inject objects that appeared in the panorama into the instructions. 
We mask out certain portions in the instructions and try to recover the missing contents with the help of the remaining instruction skeleton and the paired trajectory. 
To be specific, we use NLTK~\cite{BirdKleinLoper09} to mask out the object-related tokens in the human-annotated instructions, and the street names in the machine-generated instructions\footnote{We masked out the tokens with the following part-of-speech tags: [JJ, JJR, JJS, NN, NNS, NNP, NNPS, PDT, POS, RB, RBR, RBS, PRP\$, PRP, MD, CD]}. Multiple tokens that are masked out in a row will be replaced by a single [MASK] token. We aim to maintain the correct guiding signals for navigation after the style transfer process. Tokens that provide guiding signals, such as ``turn left" or ``take a right", will not be masked out. 
Fig.~\ref{fig:style_transfer_process} provides an example of the ``masking-and-recovering" process during training and inferring.

\noindent\textbf{Model Structure~}
Fig.~\ref{fig:style_transfer_process} illustrates the input and expected output of our multimodal text style transfer model.
We build the multimodal text style transfer model upon the Speaker model proposed by \citet{fried2018speakerfollower}. On top of the visual-attention-based LSTM~\cite{HochSchm97} structure in the Speaker model, we inject the textual attention of the masked instruction skeleton $\gX'$ to the encoder, which allows the model to attend to original guiding signals. 

The encoder takes both the visual and textual inputs, which encode the trajectory and the masked instruction skeletons. To be specific, each visual view in the trajectory is represented as a feature vector $\vv' = [\vv'_v; \vv'_{\alpha}]$, which is the concatenation of the visual encoding $\vv'_v \in \R^{ 512}$ and the orientation encoding $\vv'_\alpha\in \R^{ 64}$. The visual encoding $\vv'_v$ is the output of the last but one layer of the RESNET18~\cite{he2015deep} of the current view. The orientation encoding $\vv'_\alpha$ encodes current heading $\alpha$ by repeating vector $[sin\alpha, cos\alpha]$ for 32 times, which follows \citet{fried2018speakerfollower}. As described in section \ref{traj_encoder}, the feature matrix of a panorama is the concatenation of eight projected visual views. 

In the multimodal style transfer encoder, we use a soft-attention module~\cite{vaswani2017attention} to calculate the grounded visual feature $\hat{\vv_t}$ for current view at step $t$:
\begin{align}
    attn_{v_{t,i}} & = softmax((\mW_v\vh_{t-1})^T\vv'_i)\\ \hat{\vv_t} &= \sum_{i=1}^8= attn_{v_{t,i}}\vv'_i
\end{align}
where $\vh_{t-1}$ is the hidden context of previous step, $\mW_v$ refers to the learnable parameters, and $attn_{v_{t,i}}$ is the attention weight over the $i_{th}$ slice of view $\vv'_i$ in current panorama.

We use full-stop punctuations to split the input text into multiple sentences. 
The rationale is to enable alignment between the street views and the semantic guidance in sub-instructions.
For each sentence in the input text, the textual encoding $\vs'$ is the average of all the tokens' word embedding in the current sentence. 
We also use a soft-attention modules to calculate the grounded textual feature $\hat{\vs_t}$ at current step $t$:
\begin{align}
    attn_{s_{t,j}} & = softmax((\mW_s\vh_{t-1})^T\vs'_j) \\ \hat{\vs_t} &= \sum_{j=1}^M attn_{s_{t,j}}\vs'_j
\end{align}
where $\mW_s$ refers to the learnable parameters, $attn_{s_{t,j}}$ is the attention weight over the $j_{th}$ sentence encoding $\vs'_j$ at step $t$, and $M$ denotes the maximum sentence number in the input text. The input text for the multimodal style transfer encoder is the instruction template $\gX'$.

Based on the grounded visual feature $\hat{\vv_t}$, the grounded textual feature $\hat{\vs_t}$ and the visual view feature $\vv'_t$ at current timestamp $t$, the hidden context can be given as:
\begin{align}
    \vh_t = LSTM([\hat{\vv_t}; \hat{\vs_t}; \vv'_t])
\end{align}

\noindent\textbf{Training Objectives~}
We train the multimodal text style transfer model in the teacher-forcing manner~\cite{williams1989learning}. The decoder generates tokens auto-regressively, conditioning on the masked instruction template $\gX'$, and the trajectory.
The training objective is to minimize the following cross-entropy loss:

\small
\begin{multline}
    \mathcal{L} ( x_1,  x_2, \dots,  x_n |  \gX', \vv'_1, \dots, \vv'_N) \\
= - \log \prod_{j=1}^{n}  P(x_j | x_1, ..., x_{j-1}, \gX', \vv'_1, \dots, \vv'_N)
\end{multline}
\normalsize
where  ${x_1, x_2, \dots, x_n}$ denotes the tokens in the original instruction $\gX$, $n$ is the total token number in $\gX$, and $N$ denotes the maximum view number in the trajectory.

\subsection{VLN Transformer}

\begin{figure}[t]
\centering
\includegraphics[width=8cm]{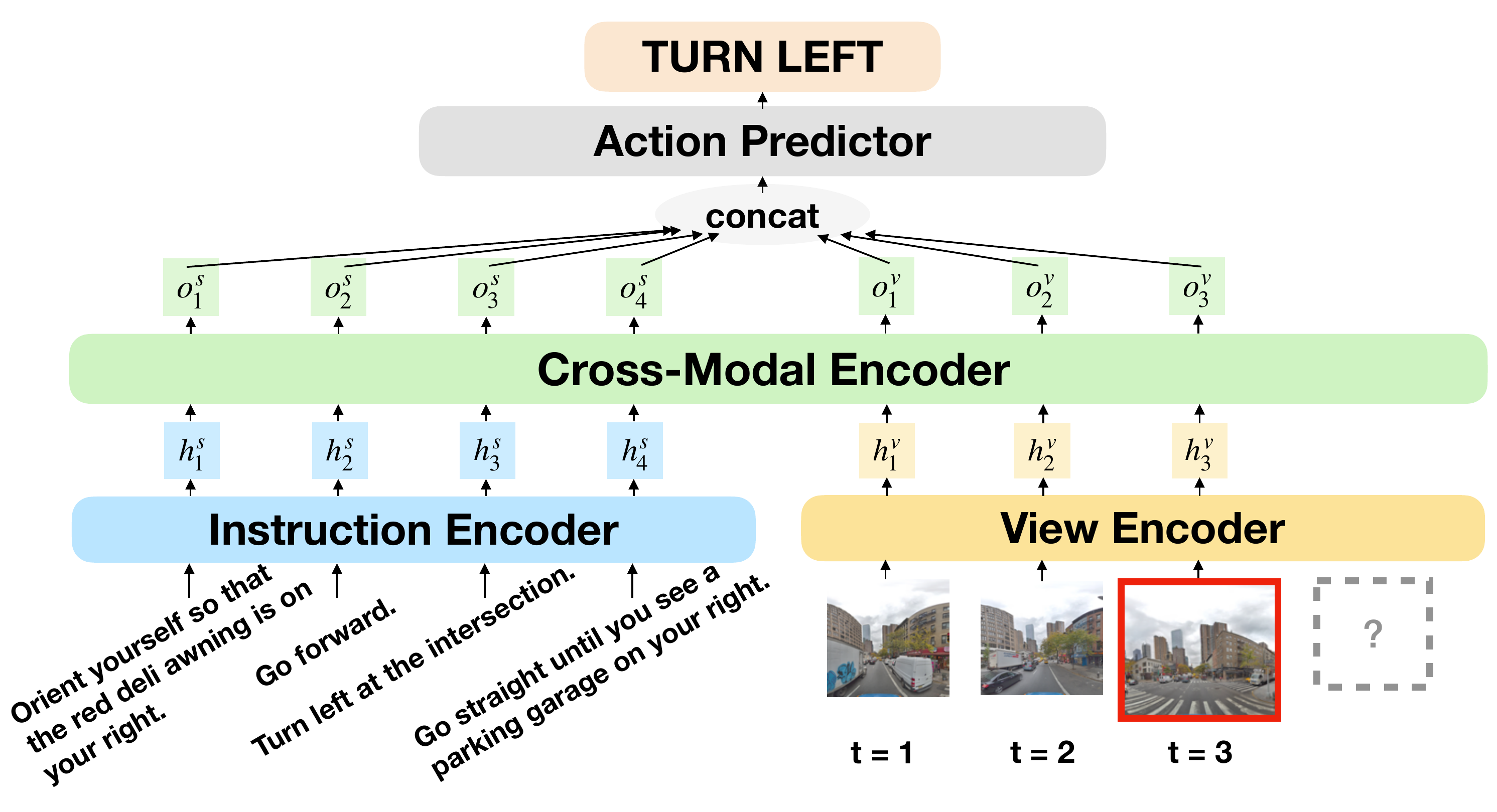}
\caption{Overview of the VLN Transformer. In this example, the VLN Transformer predicts to take a left turn for the visual scene at $t=3$.}
\label{fig:model}
\end{figure}

The VLN Transformer is the navigation agent that generates actions in the outdoor VLN task.
As illustrated in Fig.~\ref{fig:model}, our VLN Transformer is composed of an instruction encoder, a trajectory encoder, a cross-modal encoder that fuses the modality of the instruction encodings and trajectory encodings, and an action predictor. 

\noindent\textbf{Instruction Encoder~} 
\label{instr_encoder}
The instruction encoder is a pre-trained uncased BERT-base model~\cite{devlin2018bert}. 
Each piece of navigation instruction is split into multiple sentences by the full-stop punctuations. 
For the $i_{th}$ sentence $s_i = \{x_{i,1}, x_{i,2}, \dots, x_{i,l_i}\}$ that contains $l_i$ tokens, its sentence embedding $\vh_i^s$ is calculated as:
\begin{align}
    \vw_{i,j} &= \gB\gE\gR\gT(x_{i,j})  \in \R^{ 768} \\ \vh_i^s &= \gF\gC(\frac{\sum_{j=1}^{l_i}\vw_{i,j}}{l_i}) \in \R^{ 256}
\end{align}
where $\vw_{i,j}$ is the word embedding for $x_{i, j}$ generated by BERT, and $\gF\gC$ is a fully-connected layer.

\noindent\textbf{View Encoder~}
\label{traj_encoder}
We use the view encoder to retrieve embeddings for the visual views at each time step.
Following \citet{chen2019touchdown}, we embed each panorama $\rmI_t$ by slicing it into eight images and projecting each image from an equirectangular projection to a perspective projection. 
Each of the projected image of size $800 \times 460$ will be passed through the RESNET18~\cite{he2015deep} pre-trained on ImageNet~\cite{russakovsky2014imagenet}. We use the output of size $128 \times 100 \times 58$ from the fourth to last layer before classification as the feature for each slice. The feature map for each panorama is the concatenation of the eight image slices, which is a single tensor of size $128 \times 100 \times 464$.
We center the feature map according to the agent’s heading $\evalpha_t$ at timestamp $t$. We crop a $128 \times 100 \times 100$ sized feature map from the center and calculate the mean value along the channel dimension. The resulting $100 \times 100$ features is regarded as the current panorama feature $\hat{\rmI_t}$ for each state. Following \citet{mirowski2018learning}, we then apply a three-layer convolutional neural network on $\hat{\rmI_t}$ to extract the view features $\vh_t^v \in \R^{ 256}$ at timestamp $t$.

\noindent\textbf{Cross-Modal Encoder~}
In order to navigate through complicated real-world environments, the agent needs to grasp a proper understanding of the natural language instructions and the visual views jointly to choose proper actions for each state. Since the instructions and the trajectory lies in different modalities and are encoded separately, we introduce the cross-modal encoder to fuse the features from different modalities and jointly encode the instructions and the trajectory.
The cross-modal encoder is an 8-layer Transformer encoder~\cite{vaswani2017attention} with mask. We use eight self-attention heads and a hidden size of 256.

In the teacher-forcing training process, we add a mask when calculating the multi-head self-attention across different modalities. By masking out all the future views in the ground-truth trajectory, the current view $\vv_t$ is only allowed to refer to the full instructions and all the previous views that the agent has passed by, which is $[\vh_1^s, \vh_2^s, \dots, \vh_M^s; \vh_1^v, \vh_2^v, \dots, \vh_{t-1}^v]$, where $M$ denotes the maximum sentence number.

Since the Transformer architecture is based solely on attention mechanism and thus contains no recurrence or convolution, we need to inject additional information about the relative or absolute position of the features in the input sequence. We add a learned segment embedding to every input feature vector specifying whether it belongs to the sentence encodings or the view encodings. We also add a learned position embedding to indicate the relative position of the sentences in the instruction sequence or the trajectory sequence's views. 

\paragraph{Action Predictor}
The action predictor is a fully-connected layer. It takes the concatenation of the cross-modal encoder's output up to the current timestamp $t$ as input, and predicts the action $a_t$ for view $\vv_t$: 
\begin{align}
\vh_{concat} &= \vh_1^s || \dots||\vh_M^s|| \vh_1^v|| \dots|| \vh_{t}^v \\
a_t &= argmax(\gF\gC(\gT(\vh_{concat})))  
\end{align}
where $\gF\gC$ is a fully-connected layer in the action predictor, and $\gT$ refers to the Transformer operation in the cross-modal encoder. 
During training, we use the cross-entropy loss for optimization.

\section{Experiments}

\subsection{Datasets}
\paragraph{Outdoor VLN Dataset}
For the outdoor VLN task, we conduct experiments on the Touchdown dataset~\cite{chen2019touchdown,mehta2020retouchdown}, which is designed for navigation in realistic urban environments. Based on Google Street View\footnote{\url{https://developers.google.com/maps/documentation/streetview/intro}}, Touchdown's navigation environment encompasses 29,641 Street View panoramas of the Manhattan area in New York City, which are connected by 61,319 undirected edges.
The dataset contains 9,326 trajectories for the navigation task, and each trajectory is paired with a human-written instruction. The training set consists of 6,526 samples, while the development set and the test set are made up of 1,391 and 1,409 samples, respectively.

\paragraph{External Resource}
We use the StreetLearn dataset as the external resource for the outdoor VLN task~\cite{mirowski2018learning}.
The StreetLearn dataset is another dataset for navigation in real-life urban environments based on Google Street View. StreetLearn contains 114k panoramas from New York City and Pittsburgh. In the StreetLearn navigation environment, the graph for New York City contains 56k nodes and 115k edges, while the graph for Pittsburgh contains 57k nodes and 118k edges. The StreetLearn dataset contains 580k samples in the Manhattan area and 8k samples in the Pittsburgh area for navigation.

While the StreetLearn dataset's trajectory contains more panorama along the way on average, the paired instructions are shorter than the Touchdown dataset. We extract a sub-dataset \textit{Manh-50} from the original large scale StreetLearn dataset for the convenience of conducting experiments. Manh-50 consists of navigation samples in the Manhattan area that contains no more than 50 panoramas in the whole trajectory, containing 31k training samples. 
We generate style-transferred instructions for the Manh-50 dataset, which serves as an auxiliary dataset, and will be used to pre-train the navigation models. 
More details can be found in the appendix.

\subsection{Evaluation Metrics} 
We use the following metrics to evaluate VLN performance: 
(1) \emph{Task Completion (TC)}: the accuracy of completing the navigation task correctly. Following \citet{chen2019touchdown}, the navigation result is considered correct if the agent reaches the specific goal or one of the adjacent nodes in the environment graph.
(2) \emph{Shortest-Path Distance (SPD)}: the mean distance between the agent's final position and the goal position in the environment graph.
(3) \emph{Success weighted by Edit Distance (SED)}: the normalized Levenshtein edit distance between the path predicted by the agent and the reference path, which is constrained only to the successful navigation.
(4) \emph{Coverage weighted by Length Score (CLS)}: a measurement of the fidelity of the agent’s path with regard to the reference path.
(5) \emph{Normalized Dynamic Time Warping (nDTW)}: the minimized cumulative distance between the predicted path and the reference path, normalized by the reciprocal of the square root of the reference path length. The value is rescaled by taking the negative exponential of the normalized value.
(6) \emph{Success weighted Dynamic Time Warping (SDTW)}: the nDTW value where the summation is only over the successful navigation.

TC, SPD, and SED are defined by \citet{chen2019touchdown}. CLS is defined by \citet{Jain_2019}. nDTW and SDTW are originally defined by \citet{ilharco2019general}, in which nDTW is normalized by the length of the reference path. We adjust the normalizing factor to be the reciprocal of the square root of the reference path length for length invariance \citep{mueen2016extracting}.
In case the reference trajectories length has a salient variance, our modification to the normalizing factor made the nDTW and SDTW scores invariant to the reference length.


\begin{table*}[htbp]
\begin{adjustbox}{width=\linewidth,center}
\begin{tabular}{l | r r r r r r | r r r r r r}
\cmidrule[\heavyrulewidth]{1-13}
\multirow{2}{*}{Model}      & \multicolumn{6}{c}{Dev Set} & \multicolumn{6}{|c}{Test Set} \\ \cmidrule{2-13}
                            & TC $\uparrow$   & SPD $\downarrow$  & SED $\uparrow$  & CLS $\uparrow$ & nDTW $\uparrow$    & SDTW $\uparrow$ & TC $\uparrow$   & SPD $\downarrow$  & SED $\uparrow$  & CLS $\uparrow$  & nDTW $\uparrow$ & SDTW $\uparrow$ \\ \cmidrule[\heavyrulewidth]{1-13}
RCONCAT                     & 10.6  & 20.4  & 10.3  & 48.1 &  22.5 & 9.8   & 11.8  & 20.4  & 11.5 & 47.9 & 22.9  & 11.1 \\ 
+\textit{M-50}              & 11.8  & 19.1  & 11.4    & 48.7 & 23.1 & 10.9  & 12.1  & 19.4  & 11.8  & 49.4  & 24.0   & 11.3   \\ 
+\textit{M-50} +style       & 11.9  & 19.9  & 11.5  & 48.9 & 23.8 & 11.1 & 12.6  & 20.4  & 12.3  &  48.0     &  23.9 & 11.8  \\ 
\cmidrule[\heavyrulewidth]{1-13}
GA                          & 12.0  & 18.7  & 11.6  & 51.9 & 25.2 & 11.1 & 11.9 & 19.0 & 11.5 & 51.6 & 24.9 & 10.9 \\ 
+\textit{M-50}              & 12.3  & \textbf{18.5}  & 11.8  & \textbf{53.7} & 26.2 & 11.3 & 13.1 & \textbf{18.4} & 12.8 & \textbf{54.2} & 26.8 & 12.1   \\ 
+\textit{M-50} +style       & 12.9  & \textbf{18.5}  & 12.5  & 52.8 & 26.3 & 11.9 & 13.9 & \textbf{18.4} & 13.5 & 53.5 & 27.5 & 12.9   \\ 
\cmidrule[\heavyrulewidth]{1-13}
VLN Transformer             & 14.0  & 21.5  & 13.6  & 44.0 & 23.0      & 12.9  & 14.9  & 21.2  & 14.6  & 45.4 & 25.3  & 14.0 \\ 
+\textit{M-50}              & 14.6  & 22.3  & 14.1  & 45.6 & 25.0      & 13.4  & 15.5  & 21.9  & 15.4  & 45.9 & 26.1  & 14.2 \\ 
+\textit{M-50} +style       & \textbf{15.0}  & 20.3  & \textbf{14.7}  & 50.1 & \textbf{27.0} & \textbf{14.2}  & \textbf{16.2}  & 20.8  & \textbf{15.7}  & 50.5 & \textbf{27.8}  & \textbf{15.0} \\ 
\cmidrule[\heavyrulewidth]{1-13}
\end{tabular}
\end{adjustbox}
\caption{Navigation results on the outdoor VLN task. +\textit{M-50} denotes pre-training with vanilla Manh-50 which contains machine-generated instructions;
in the \textit{+style} setting, the model is pre-trained with Manh-50 trajectories and style-modified instructions that are generated by our MTST model.}
\label{tab:vln_result}
\end{table*}

\begin{table*}[htbp]
\setlength{\tabcolsep}{4pt}
\begin{adjustbox}{width=\linewidth,center}
\begin{tabular}{l | r r r r r r | r r r r r r}
\cmidrule[\heavyrulewidth]{1-13}
\multirow{2}{*}{Model}      & \multicolumn{6}{c}{Dev Set} & \multicolumn{6}{|c}{Test Set} \\ \cmidrule{2-13}
                            & TC $\uparrow$   & SPD $\downarrow$  & SED $\uparrow$  & CLS $\uparrow$ & nDTW $\uparrow$    & SDTW $\uparrow$ & TC $\uparrow$   & SPD $\downarrow$  & SED $\uparrow$  & CLS $\uparrow$  & nDTW $\uparrow$ & SDTW $\uparrow$ \\ \cmidrule[\heavyrulewidth]{1-13}
VLN Transformer +\textit{M-50}& 14.6  & 22.3  & 14.1  & 45.6 & 25.0      & 13.4  & 15.5  & 21.9  & 15.4  & 45.9 & 26.1  & 14.2  \\ 
+speaker                    & 7.6 & 26.2  & 7.3  & 34.6 & 14.6      & 7.0 & 8.3  & 25.4 & 8.0   & 36.3   & 15.9 & 7.7  \\ 
+text\_attn                 & 11.7 & \textbf{20.1} & 11.3 & 46.3  &  23.2     & 10.7  & 11.8  & \textbf{20.5} & 11.5  &  47.3 & 23.2 & 11.0  \\ 
+style                      & \textbf{15.0}  & 20.3  & \textbf{14.7}  & \textbf{50.1} & \textbf{27.0} & \textbf{14.2}  & \textbf{16.2}  & 20.8  & \textbf{15.7}  & \textbf{50.5} & \textbf{27.8}  & \textbf{15.0} \\ 
\cmidrule[\heavyrulewidth]{1-13}
\end{tabular}
\end{adjustbox}
\caption{Ablation study of the multimodal text style transfer model on the outdoor VLN task.
In the \textit{+speaker} setting, the instructions used in pre-training are generated by the Speaker~\cite{fried2018speakerfollower}, which only attends to the visual input;
\textit{+text\_attn} denotes that we add a textual attention module to the Speaker to attend to both the visual input and the machine-generated instructions provided by Google Maps API.
}
\label{tab:style_transfer_components}
\end{table*}

\begin{table}[htbp]
\begin{adjustbox}{width=\linewidth,center}
\begin{tabular}{l | r r | r r}
\cmidrule[\heavyrulewidth]{1-5}
\multirow{2}{*}{Model}      & \multicolumn{2}{c}{Dev Set} & \multicolumn{2}{|c}{Test Set} \\ \cmidrule{2-5}
      & S\_SPD$\downarrow$ & F\_SPD$\downarrow$ &  S\_SPD$\downarrow$ & F\_SPD$\downarrow$ \\ \cmidrule[\heavyrulewidth]{1-5}
RCONCAT                      & 0.64 & 22.68  & 0.67 & 23.06 \\
+\textit{M-50}               & 0.68 & 21.53  & 0.69 & 21.97 \\
+\textit{M-50} +style        & 0.66 & 22.48  & 0.69 & 23.21 \\
\cmidrule[\heavyrulewidth]{1-5}
GA                           & 0.65 & 21.15  & 0.66 & 21.41 \\
+\textit{M-50}               & 0.70 & \textbf{20.95}  & 0.77 & \textbf{21.09} \\ 
+\textit{M-50} +style        & 0.65 & 21.11  & 0.70 & 21.26 \\
\cmidrule[\heavyrulewidth]{1-5}
VLN Transformer              & 0.66 & 24.92  & 0.63 & 24.84 \\
+\textit{M-50}               & 0.67 & 25.94  & 0.63 & 25.77 \\
+\textit{M-50} +style        & \textbf{0.59} & 23.72  & \textbf{0.62} & 24.67 \\
\cmidrule[\heavyrulewidth]{1-5}
\end{tabular}
\end{adjustbox}
\caption{\textit{S\_SPD} and \textit{F\_SPD} denotes the average SPD value on success and failure cases respectively.}
\label{tab:spd_result}
\end{table}

\subsection{Results and Analysis}

In this section, we report the outdoor VLN performance and the quality of the generated instructions to validate the effectiveness of our MTST learning approach. 
We compare our VLN Transformer with the baseline model and discuss the influence of pre-training on external resources with/without instruction style transfer.

\paragraph{Outdoor VLN Performance} 
We compare our VLN Transformer with RCONCAT~\cite{chen2019touchdown,mirowski2018learning} and GA~\cite{chen2019touchdown,Chaplot2018GatedAttentionAF} as baseline models. Both baseline models encode the trajectory and the instruction in an LSTM-based manner and use supervised training with Hogwild!~\cite{niu2011hogwild}. 
Table~\ref{tab:vln_result} presents the navigation results on the Touchdown validation and test sets, where VLN Transformer performs better than RCONCAT and GA on most metrics with the exception of SPD and CLS. 

Pre-training the navigation models on Manh-50 with template-based instructions can partially improve navigation performance. For all three agent models, the scores related to successful cases---such as TC, SED, and SDTW---witness a boost after being pre-trained on vanilla Manh-50. 
However, the instruction style difference between Manh-50 and Touchdown might misguide the agent in the pre-training stage, resulting in a performance drop on SPD for our VLN Transformer model. 

In contrast, our MTST learning approach can better utilize external resources and further improve navigation performance. Pre-training on Manh-50 with style-modified instructions can stably improve the navigation performance on all the metrics for both the RCONCAT model and the VLN Transformer. This also indicates that our MTST learning approach is model-agnostic.

Table~\ref{tab:spd_result} compares the SPD values on success and failure navigation cases. 
In the success cases, VLN Transformer has better SPD scores, which is aligned with the best SED results in Table~\ref{tab:vln_result}. Our model's inferior SPD results are caused by taking longer paths in failure cases, which also harms the fidelity of the generated path and lowers the CLS scores. Nevertheless, every coin has two sides, and exploring more areas when getting lost might not be a complete bad behavior for the navigation agent. We leave this to future study.

\paragraph{Multimodal Text Style Transfer in VLN}
We attempt to reveal each component's effect in the multimodal text style transfer model. We pre-train the VLN Transformer with external trajectories and instructions generated by different models, then fine-tune it on the TouchDown dataset.

According to the navigation results in Table~\ref{tab:style_transfer_components}, the instructions generated by the Speaker model misguide the navigation agent, indicating that relying solely on the Speaker model cannot reduce the gap between different instruction styles. Adding textual attention to the Speaker model can slightly improve the navigation results, but still hinders the agent from navigating correctly. The style-modified instructions improve the agent's performance on all the navigation metrics, suggesting that our Multimodal Text Style Transfer learning approach can assist the outdoor VLN task.

\paragraph{Quality of the Generated Instruction}
We evaluate the quality of instructions generated by the Speaker and the MTST model.
We utilize five automatic metrics for natural language generation to evaluate the quality of the generated instructions, including BLEU~\cite{bleu}, ROUGE~\cite{rouge}, METEOR~\cite{meteor}, CIDEr~\cite{cider} and SPICE~\cite{Anderson2016SPICESP}. 
In addition, we calculate the guiding signal match rate (MR) by comparing the appearance of ``turn left'' and ``turn right''. If the generated instruction contains the same number of guiding signals in the same order as the ground truth instruction, then this instruction pair is considered to be matched.
We also calculate the number of different infilled tokens (\#infill) in the generated instruction\footnote{We regard tokens with the following part-of-speech tags as infilled tokens: [JJ, JJR, JJS, NN, NNS, NNP, NNPS, PDT, POS, RB, RBR, RBS, PRP\$, PRP, MD, CD]}. This reflects the model's ability to inject object-related information during style transferring.  
Among the 9,326 trajectories in the Touchdown dataset, 9,000 are used to train the MTST model, while the rest form the validation set.

We report the quantitative results on the validation set in Table~\ref{tab:nlg_metrics}.
After adding textual attention to the Speaker, the evaluation performance on all seven metrics improved. Our MTST model scores the highest on all seven metrics, which indicates that the ``masking-and-recovering'' scheme is beneficial for the multimodal text style transfer process. The results validate that the MTST model can generate higher quality instructions, which refers to more visual objects and provide more matched guiding signals.

\begin{table}[t]
\setlength{\tabcolsep}{2.5pt}
\begin{adjustbox}{width=\linewidth,center}

\begin{tabular}{l | c c c c c c c}
\cmidrule[\heavyrulewidth]{1-8}
Model       & BLEU  & METEOR    & ROUGE\_L  & CIDEr & SPICE & MR & \#infill  \\ \cmidrule[\heavyrulewidth]{1-8}
Speaker    & 15.1  & 20.6      & 22.2      & 1.4   & 20.7  & 8.3 &  160 \\ 
Text\_Attn & 23.8  & 23.3      & 29.6      & 10.0  & 24.6  & 35.7 &  182 \\ 
MTST      & \textbf{30.6}  & \textbf{28.8}      & \textbf{39.7}      & \textbf{27.8} & \textbf{30.6}  & \textbf{46.7} &  \textbf{308} \\
\cmidrule[\heavyrulewidth]{1-8}
\end{tabular}
\end{adjustbox}
\caption{Quantitative evaluation of the instructions generated by Speaker, Speaker with textual attention and our MTST model.}
\label{tab:nlg_metrics}
\end{table}


\begin{table*}[htbp]
\setlength{\tabcolsep}{4pt}
\begin{adjustbox}{width=0.9\linewidth,center}
\begin{tabular}{l | r r r | r r r | r r r }
\cmidrule[\heavyrulewidth]{1-10}
\multirow{2}{*}{Choice (\%)}      & \multicolumn{3}{c}{MTST \textit{vs} Speaker} & \multicolumn{3}{|c}{MTST \textit{vs} Text\_Attn} & \multicolumn{3}{|c}{Speaker \textit{vs} Text\_Attn} \\ \cmidrule{2-10}
                            & MTST & Speaker & Tie & MTST & Text\_Attn & Tie & Speaker & Text\_Attn & Tie \\ \cmidrule[\heavyrulewidth]{1-10}
Better describes the street view  & \textbf{67.9}& 22.8  & 9.3       & \textbf{44.3} & 35.8 & 19.9  & 28.2  & \textbf{62.7}  & 9.1    \\ 
More aligned with the ground truth & \textbf{64.6}  & 26.8  & 8.6   & \textbf{37.6} & 33.9 & 28.5  & 25.3  & \textbf{62.5}  & 12.2    \\ 
\cmidrule[\heavyrulewidth]{1-10}
\end{tabular}
\end{adjustbox}
\caption{Human evaluation results of the instructions generated by Speaker, Speaker with textual attention and our MTST model with pairwise comparisons.}
\label{tab:mturk}
\end{table*}

\paragraph{Human Evaluation}
We invite human judges on Amazon Mechanical Turk to evaluate the quality of the instructions generated by different models.
We conduct a pairwise comparison, which covers 170 pairs of instructions generated by Speaker, Speaker with textual attention, and our MTST model. The instruction pairs are sampled from the Touchdown validation set. 
Each pair of instructions, together with the ground truth instruction and the gif that illustrates the navigation street view, is presented to 5 annotators. The annotators are asked to make decisions from the aspect of guiding signal correctness and instruction content alignment.
Results in Table~\ref{tab:mturk} show that annotators think the instructions generated by our MTST model better describe the street view and is more aligned with the ground-truth instructions.

\begin{figure}[t]
\vspace{-3ex}
\centering
\includegraphics[width=\linewidth]{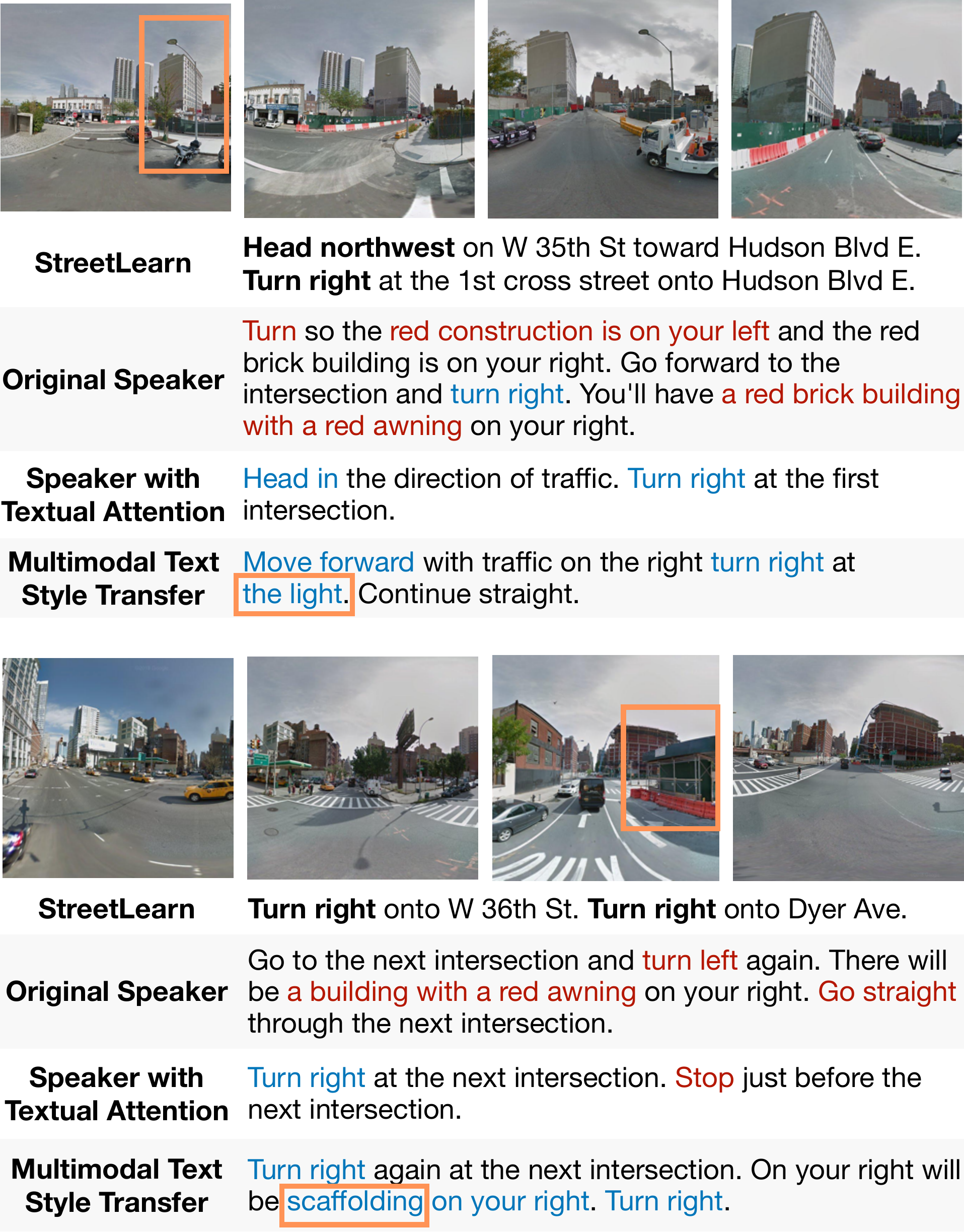}
\caption{Two showcases of the instruction generation results. 
The red tokens indicate incorrectly generated instructions, while the blue tokens suggest alignments with the ground truth. The orange bounding boxes show that the objects in the surrounding environment have been successfully injected into the style-modified instruction. }
\label{fig:case_study_style}
\vspace{-2ex}
\end{figure}

\paragraph{Case Study}
We demonstrate case study results to illustrate the performance of our Multimodal Text Style Transfer learning approach. Fig.~\ref{fig:case_study_style} provides two showcases of the instruction generation results. As listed in the charts, the instructions generated by the vanilla Speaker model have a poor performance in keeping the guiding signals in the ground truth instructions and suffer from hallucinations, which refers to objects that have not appeared in the trajectory. 
The Speaker with textual attention can provide guidance direction. However, the instructions generated in this manner does not utilize the rich visual information in the trajectory. On the other hand, the instructions generated by our multimodal text style transfer model inject more object-related information (``the light", ``scaffolding") in the surrounding navigation environment to the StreetLearn instruction while keeping the correct guiding signals.

\section{Conclusion}
In this paper, we proposed the Multimodal Text Style Transfer learning approach for outdoor VLN. This learning framework allows us to utilize out-of-domain navigation samples in outdoor environments and enrich the original navigation reasoning training process. Experimental results show that our MTST approach is model-agnostic, and our MTST learning approach outperforms the baseline models on the outdoor VLN task.
We believe our study provides a possible solution to mitigate the data scarcity issue in the outdoor VLN task. In future studies, we would love to explore the possibility of constructing an end-to-end framework. We will also further improve the quality of style-modified instructions,  and quantitatively evaluate the alignment between the trajectory and the style-transferred instructions. 

\section*{Acknowledgments}
We would like to show our gratitude towards Jiannan Xiang, who kindly shares his experimental code on Touchdown, and Qi Wu, who provides valuable feedback to our initial draft. We also thank the anonymous reviewers for their thought-provoking comments.
The UCSB authors were sponsored by an unrestricted gift from Google. The views and conclusions contained in this document are those of the authors and should not be interpreted as representing the sponsor.

\bibliography{anthology,eacl2021}
\bibliographystyle{acl_natbib}

\clearpage

\appendix

\section{Appendix}

\subsection{Dataset Comparison~}

\begin{table}[h]
\setlength{\tabcolsep}{3pt}
\small
\begin{adjustbox}{width=\linewidth,center}
\begin{tabular}{l r r r r r r}
\cmidrule[\heavyrulewidth]{1-7}
\textbf{Dataset} & \textbf{\#path}   & \textbf{\#pano} &  \textbf{\#pano/path}  & \textbf{instr\_len} & \textbf{\#sent/path} & \textbf{\#turn/path} \\ \cmidrule{1-7}
Touchdown        & 6k  & 26k & 35.2 & 80.5 & 6.3 & 2.8  \\ \cmidrule{1-7}
Manh-50         & 31k  & 43k & 37.2 & 22.1 & 2.8 & 4.1  \\ \cmidrule{1-7}
StreetLearn      & 580k   & 114k & 29.0    & 28.6  & 4.0    & 13.2  \\ 
\cmidrule[\heavyrulewidth]{1-7}
\end{tabular}
\end{adjustbox}
\caption{Dataset statistics. \emph{path}: navigation path; \emph{pano}: panorama; \emph{instr\_len}: average instruction length; \emph{sent}: sentence; \emph{turn}: intersection on the path. 
}
\label{tab:dataset_details}
\end{table}

 Table \ref{tab:dataset_details} lists out the statistical information of the datasets used in pre-training and fine-tuning.
Even though the Touchdown dataset and the StreetLearn dataset are built upon Google Street View, and both of them contain urban environments in New York City, pre-training the model with the VLN task on the StreetLearn dataset does not raise a threat of test data leaking. This is due to several causes: 

First, the instructions in the two datasets are distinct in styles. The instructions in the StreetLearn dataset is generated by Google Maps API, which is template-based and focuses on street names. However, the instructions in the Touchdown dataset are created by human annotators and emphasize the visual environment's attributes as navigational cues. Moreover, as reported by \citet{mehta2020retouchdown}, the panoramas in the two datasets have little overlaps. In addition, Touchdown instructions constantly refer to transient objects such as cars and bikes, which might not appear in a panorama from a different time. The different granularity of the panorama spacing also leads to distinct panorama distributions of the two datasets.

\subsection{Training Details}

We use Adam optimizer~\cite{kingma2014adam} to optimize all the parameters. During pre-training on the StreetLearn dataset, the learning rate for the RCONCAT model, GA model, and the VLN Transformer is $2.5 \times 10^{-4}$. We fine-tune BERT separately with a learning rate of $1 \times 10^{-5}$. We pre-train RCONCAT and GA for 15 epochs and pre-train the VLN Transformer for 25 epochs.

When training or fine-tuning on the Touchdown dataset, the learning rate for RCONCAT and GA is $2.5 \times 10^{-4}$. For the VLN Transformer, the learning rate to fine-tune BERT is initially set to $1 \times 10^{-5}$, while the learning rate for other parameters in the model is initialized to be $2.5 \times 10^{-4}$. The learning rate for VLN Transformer will decay.
The batch size for RCONCAT and GA is 64, while the VLN Transformer uses a batch size of 30 during training.

\begin{table}[h]
\begin{adjustbox}{width=\linewidth,center}
\begin{tabular}{l | r r r r r r }
\cmidrule[\heavyrulewidth]{1-7}
Model & TC $\uparrow$   & SPD $\downarrow$  & SED $\uparrow$  & CLS $\uparrow$ & nDTW $\uparrow$    & SDTW $\uparrow$ \\ \cmidrule[\heavyrulewidth]{1-7}
no split & 9.6  & 21.8  & 9.3  & 46.1 & 20.0      &  8.7  \\ 
split & \textbf{13.6}  & \textbf{20.5}  & \textbf{13.1}  & \textbf{47.6} & \textbf{24.0}      & \textbf{12.6}   \\ 
\cmidrule[\heavyrulewidth]{1-7}
\end{tabular}
\end{adjustbox}
\caption{Ablation results of the VLN Transformer's instruction split on Touchdown dev set. In \textit{split} setting, the instruction is split into multiple sentences before being encoded by the instruction encoder, while \textit{no split} setting encodes the whole instruction without splitting.}
\label{tab:split}
\end{table}

\subsection{Split Instructions vs. No Split~}
We compare VLN Transformer performance with and without splitting the instructions into sentences during encoding. Results in Table ~\ref{tab:split} show that breaking the instructions into multiple sentences allows the visual views and the guiding signals in sub-instructions to attend to each other during cross-modal encoding fully. Such cross-modal alignments lead to betters navigation performance.

\subsection{Amazon Mechanical Turk}
\begin{figure*}[t]
\centering
\includegraphics[width=\linewidth]{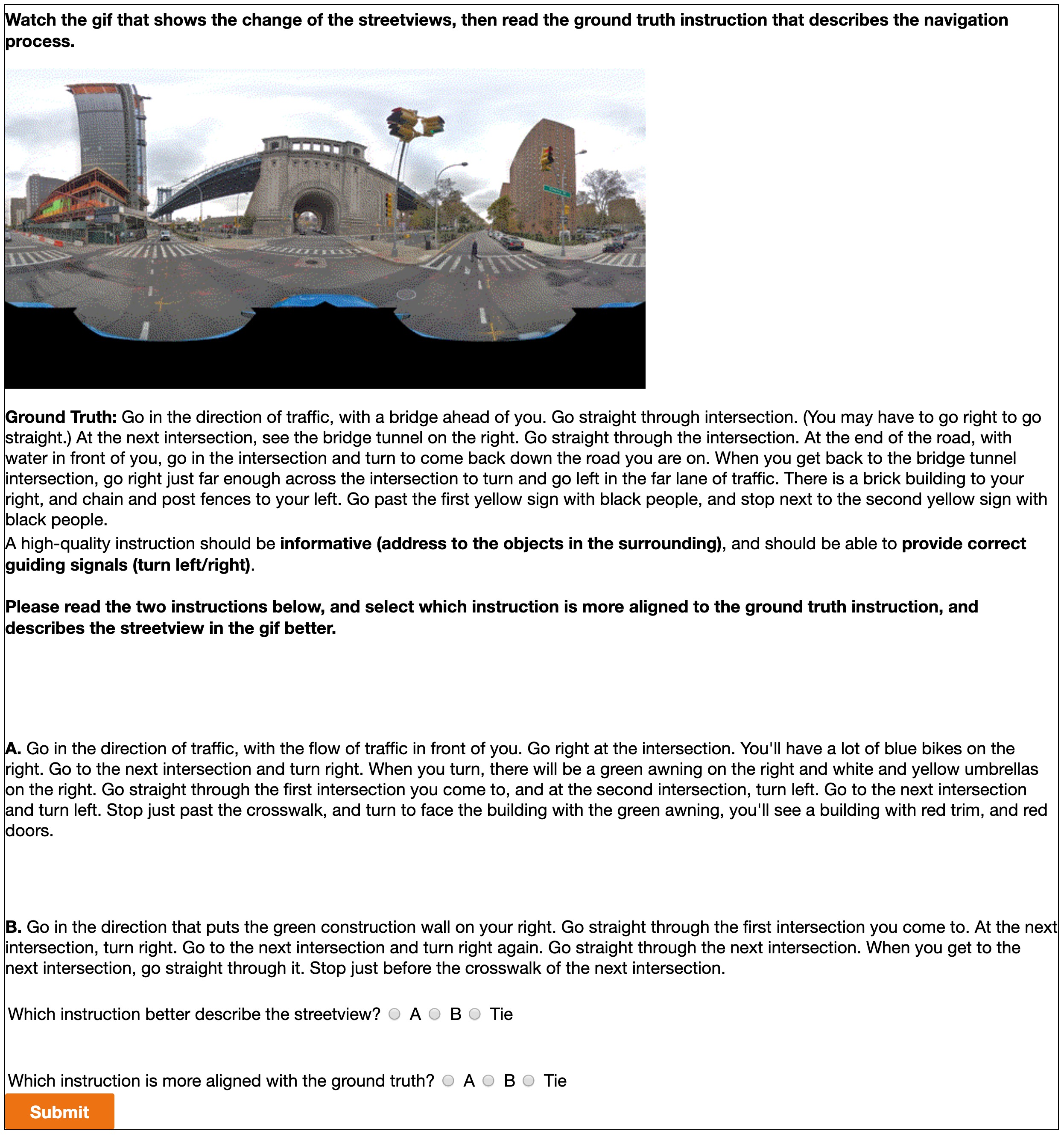}
\caption{Pairwise comparison form for human evaluation on AMT.}
\label{fig:mturk}
\end{figure*}

We use AMT for human evaluation when evaluating the quality of the instructions generated by different models. The survey form for head-to-head comparisons is shown in Figure~\ref{fig:mturk}.

\end{document}